%% file: main.tex
\def\year{2022}\relax
\documentclass[letterpaper]{article} % DO NOT CHANGE THIS
\usepackage{aaai22}  % DO NOT CHANGE THIS
\usepackage{times}  % DO NOT CHANGE THIS
\usepackage{helvet}  % DO NOT CHANGE THIS
\usepackage{courier}  % DO NOT CHANGE THIS
\usepackage[hyphens]{url}  % DO NOT CHANGE THIS
\usepackage{graphicx} % DO NOT CHANGE THIS
\usepackage{natbib}  % DO NOT CHANGE THIS AND DO NOT ADD ANY OPTIONS TO IT
\usepackage{caption} % DO NOT CHANGE THIS AND DO NOT ADD ANY OPTIONS TO IT
\DeclareCaptionStyle{ruled}{labelfont=normalfont,labelsep=colon,strut=off} % DO NOT CHANGE THIS
\usepackage{algorithm}
\usepackage{booktabs}
\usepackage{amsmath}
\usepackage{amssymb}
\usepackage{amsfonts}
\usepackage{subfigure}
\newtheorem{definition}{Definition}
\newtheorem{example}{Example}
\usepackage{algorithmicx}

\title{Logical Credal Networks}
\author {
    Haifeng Qian,\textsuperscript{\rm 1}
    Radu Marinescu,\textsuperscript{\rm 2}
    Alexander Gray,\textsuperscript{\rm 1}
    Debarun Bhattacharjya,\textsuperscript{\rm 1}
    Francisco Barahona,\textsuperscript{\rm 1}
    Tian Gao,\textsuperscript{\rm 1}
    Ryan Riegel,\textsuperscript{\rm 1}
    Pravinda Sahu\textsuperscript{\rm 3}
}
\affiliations {
    \textsuperscript{\rm 1} IBM T. J. Watson Research Center, Yorktown Heights, NY, USA\\
    \textsuperscript{\rm 2} IBM Research Europe, Dublin, Ireland\\
    \textsuperscript{\rm 3} IBM Global Business Services, Bangalore, KA, India\\
    qianhaifeng@us.ibm.com, radu.marinescu@ie.ibm.com, alexander.gray@ibm.com, debarunb@us.ibm.com, barahon@us.ibm.com, tgao@us.ibm.com, ryan.riegel@ibm.com, pravisah@in.ibm.com
}

\begin{document}

\include{macros}
\maketitle

\begin{abstract}
This paper introduces Logical Credal Networks, an expressive probabilistic logic that generalizes many prior %works
models
that combine logic and probability. Given imprecise information represented by probability bounds and conditional probability bounds of logic formulas, this logic specifies a set of probability distributions over all interpretations. On the one hand, %this logic 
our approach
allows propositional and first-order logic formulas with few restrictions, e.g., without requiring acyclicity. On the other hand, it has a Markov condition similar to Bayesian networks and Markov random fields that is critical in real-world applications. Having both these properties makes this logic unique, and we investigate its performance on maximum a posteriori inference tasks, including solving Mastermind games with uncertainty and detecting credit card fraud. The results show that the proposed method outperforms existing approaches, and its advantage lies in aggregating multiple sources of imprecise information.
\end{abstract}

\input{introduction}

\input{background}

\input{lcn}

\input{results}

\section{Conclusions}

We propose a new probabilistic logic that expresses both probability bounds for propositional and first-order logic formulas with little restrictions and a Markov condition that is similar to Bayesian networks and Markov random fields.
The formula bounds allow for flexibility in the form and precision of background knowledge that can be utilized, while the Markov condition restricts the space of distributions to enable a meaningful representation of uncertainties. 
Evaluation on a set of MAP inference tasks shows promising results, particularly in aggregating multiple sources of imprecise information.
Potential future directions include extending to temporal models, further algorithmic innovations and experiments on a wider array of applications.

\bibliography{ref}

\input{sup}

\end{document}

%% file: macros.tex
%%%%%%%%%%%%%%%%%%%%%%%%%%%%%%%%%%%%%%%%%%%%%
%%        math bold fonts                  %%
%%%%%%%%%%%%%%%%%%%%%%%%%%%%%%%%%%%%%%%%%%%%%
\newcommand{\bA}{\mathbf{A}}
\newcommand{\bB}{\mathbf{B}}
\newcommand{\bC}{\mathbf{C}}
\newcommand{\bD}{\mathbf{D}}
\newcommand{\bE}{\mathbf{E}}
\newcommand{\bF}{\mathbf{F}}
\newcommand{\bG}{\mathbf{G}}
\newcommand{\bH}{\mathbf{H}}
\newcommand{\bI}{\mathbf{I}}
\newcommand{\bJ}{\mathbf{J}}
\newcommand{\bK}{\mathbf{K}}
\newcommand{\bL}{\mathbf{L}}
\newcommand{\bM}{\mathbf{M}}
\newcommand{\bN}{\mathbf{N}}
\newcommand{\bO}{\mathbf{O}}
\newcommand{\bP}{\mathbf{P}}
\newcommand{\bQ}{\mathbf{Q}}
\newcommand{\bR}{\mathbf{R}}
\newcommand{\bS}{\mathbf{S}}
\newcommand{\bT}{\mathbf{T}}
\newcommand{\bU}{\mathbf{U}}
\newcommand{\bV}{\mathbf{V}}
\newcommand{\bW}{\mathbf{W}}
\newcommand{\bX}{\mathbf{X}}
\newcommand{\bY}{\mathbf{Y}}
\newcommand{\bZ}{\mathbf{Z}}

\newcommand{\ba}{\mathbf{a}}
\newcommand{\bb}{\mathbf{b}}
\newcommand{\bc}{\mathbf{c}}
\newcommand{\bd}{\mathbf{d}}
\newcommand{\be}{\mathbf{e}}
\newcommand{\bbf}{\mathbf{f}}
\newcommand{\bg}{\mathbf{g}}
\newcommand{\bh}{\mathbf{h}}
\newcommand{\bi}{\mathbf{i}}
\newcommand{\bj}{\mathbf{j}}
\newcommand{\bk}{\mathbf{k}}
\newcommand{\bl}{\mathbf{l}}
\newcommand{\bm}{\mathbf{m}}
\newcommand{\bn}{\mathbf{n}}
\newcommand{\bo}{\mathbf{o}}
\newcommand{\bp}{\mathbf{p}}
\newcommand{\bq}{\mathbf{q}}
\newcommand{\br}{\mathbf{r}}
\newcommand{\bs}{\mathbf{s}}
\newcommand{\by}{\mathbf{y}}
\newcommand{\bu}{\mathbf{u}}
\newcommand{\bv}{\mathbf{v}}
\newcommand{\bw}{\mathbf{w}}
\newcommand{\bx}{\mathbf{x}}
\newcommand{\bby}{\mathbf{y}}
\newcommand{\bz}{\mathbf{z}}

%%%%%%%%%%%%%%%%%%%%%%%%%%%%%%%%%%%%%%%%%%%%%
%%        math cal fonts                  %%
%%%%%%%%%%%%%%%%%%%%%%%%%%%%%%%%%%%%%%%%%%%%%
\newcommand{\cA}{\mathcal{A}}
\newcommand{\cB}{\mathcal{B}}
\newcommand{\cC}{\mathcal{C}}
\newcommand{\cD}{\mathcal{D}}
\newcommand{\cE}{\mathcal{E}}
\newcommand{\cF}{\mathcal{F}}
\newcommand{\cG}{\mathcal{G}}
\newcommand{\cH}{\mathcal{H}}
\newcommand{\cI}{\mathcal{I}}
\newcommand{\cJ}{\mathcal{J}}
\newcommand{\cK}{\mathcal{K}}
\newcommand{\cL}{\mathcal{L}}
\newcommand{\cM}{\mathcal{M}}
\newcommand{\cN}{\mathcal{N}}
\newcommand{\cO}{\mathcal{O}}
\newcommand{\cP}{\mathcal{P}}
\newcommand{\cQ}{\mathcal{Q}}
\newcommand{\cR}{\mathcal{R}}
\newcommand{\cS}{\mathcal{S}}
\newcommand{\cT}{\mathcal{T}}
\newcommand{\cU}{\mathcal{U}}
\newcommand{\cV}{\mathcal{V}}
\newcommand{\cW}{\mathcal{W}}
\newcommand{\cX}{\mathcal{X}}
\newcommand{\cY}{\mathcal{Y}}
\newcommand{\cZ}{\mathcal{Z}}

% \inmath{material}
%    Make sure that {material} is inside math mode.
\def\inmath#1{\relax\ifmmode#1\else$#1$\fi}

% \@<token>
%    Will print the argument in \cal.
\def\@#1{\inmath{{\cal #1}}}

% \operators{<list>}
% \variables{<list>}
%    A convenient way of defining a set of variables or functions for
%    use in math. <list> consists of a \\ separated list of pairs
%    <token><text> where <token> is the token that prints <text> as an
%    operator/variable in math-mode (using \mathop or \mathord).
%
%    See \det and \gcd below for examples.
%
\outer\def\operators#1{\begingroup
        \def\\##1##2{\gdef##1{{\mathop{\rm ##2}}}}\relax
        \\#1\endgroup
}

\outer\def\variables#1{\begingroup
        \def\\##1##2{\gdef##1{{\mathord{\it ##2}}}}\relax
        \\#1\endgroup
}

% \ntimes#1#2
% make #1 repetitions of #2
\def\ntimes#1#2{{\count255=1\loop#2\ifnum\count255<#1\advance\count255 by1\repeat}}

% \boxit{material}
%    This one boxes {material} using rules and is a vbox itself
\long\def\boxit#1{\vbox{\hrule\hbox{\vrule\kern3pt\vbox{\kern3pt#1\kern3pt}\kern3pt\vrule}\hrule}}

\long\def\entry#1#2\par{\goodbreak\item{$\bullet$}\strut{#1}\par\nobreak{\narrower\vbox{#2\strut}}\par}

% \th{1}
%    Adds apropriate ``th'' extension to digits (``1st'', ``2nd'', ``3d'', ....)
\def\th#1{#1\ifcase#1 th\or st\or nd\or d\else th\fi}

% \smiley
% \frowny
%    Here we have a smile face (and a frowny face) to use together with
%    text in horisontal mode
\begingroup
        \def\facewith#1{$\bigcirc\mskip-13.3mu{}^{..}
        \mskip-11mu\scriptscriptstyle#1\ $}
        \xdef\frowny{\facewith\frown}
        \xdef\smiley{\facewith\smile}
\endgroup

% \underrightarrow{material}
% \underleftarrow{material}
%    Insert a right or left arrow under {material}, does not affect the baseline
\def\underrightarrow#1{\vtop{\ialign{##\crcr
        \hbox{#1}\crcr
        \noalign{\kern-1pt\nointerlineskip}
        \rightarrowfill\crcr}}}
\def\underleftarrow#1{\vtop{\ialign{##\crcr
        \hbox{#1}\crcr
        \noalign{\kern-1pt\nointerlineskip}
        \leftarrowfill\crcr}}}

% \addtobox\foo{bar}
%    This one adds {bar} to the box represented by \foo.
%    If the box is void, the material is inserted into the box
%
%    NOTE! The box \foo has to be of the same type as the {bar} material
\def\addtobox#1#2{\setbox#1=\ifvoid#1
  {#2}
  \else\ifvbox#1
    \vbox{\unvbox#1\relax#2}
    \else\ifhbox#1
      \hbox{\unhbox#1\relax#2}
    \fi\fi\fi}

% \today
%    This one constructs todays date in the format ``November 19, 1999''
\def\today{\ifcase\month\or January\or February\or March\or April\or May\or June\or July\or
        August\or September\or October\or November\or December\fi
        \space\number\day, \number\year}

% \idag
%    A swedish version of the \today macro.
\def\idag{\number\day\space
        \ifcase\month\or Januari\or Februari\or Mars\or April\or Maj\or Juni\or Juli\or
        Augusti\or September\or Oktober\or November\or December\fi
        \space\number\year}

% \mwidthof{foo}
% \mheightof{foo}
%    Prints the height or the width of {foo} as a message
\def\mwidthof#1{\setbox0=\hbox{#1}\message{\the\wd0}}
\def\mheightof#1{\setbox0=\hbox{#1}\message{\the\ht0}}

% \widthof{foo}
% \heightof{foo}
%    Expands to the width or the height of {foo}.
\def\widthof#1{\setbox0=\hbox{#1}\the\wd0}
\def\heightof#1{\setbox0=\hbox{#1}\the\ht0}

% \nextpar{sequence}
%    Will enter {sequence} at the beginning of the next paragraph.
\def\nextpar#1{\everypar={#1\everypar={}}}

% \dropnextpar
%    Will enlarge the first letter of the next paragraph and drop it
%    into the following text in the manner that some storybooks (and
%    old bibles) do
\def\dropnextpar{\begingroup\parindent=0pt
        \font\bigfont=cmr10 at 2\baselineskip
        \everypar={\futurelet\a\cont}}
\def\cont#1#2{\setbox0=\hbox{\bigfont#1\kern1.5pt}\relax
        \setbox1=\hbox{#2}\hangindent=\wd0
        \hangafter=-\ht0\advance\hangafter by -\baselineskip
        \divide\hangafter by\baselineskip
        \noindent\llap{\vbox to\the\ht1{\vskip-.5pt\box0\vss}}#2\endgroup}

% \registered
%    Registered trademark symbols
\def\registered{\raise.5ex\hbox{\ooalign{\hfil\raise.06ex
        \hbox{$\scriptstyle\rm R$}\hfil\crcr\mathhexbox20D}}}

%%% Local Variables: 
%%% mode: latex
%%% TeX-master: t
%%% End: 

%% file: introduction.tex
\section{Introduction}
\label{sec-intro}

Probabilistic graphical models \cite{geman1984,pearl88,mln,koller2009probabilistic} are a powerful formalism for reasoning under
uncertainty. However, these models typically represent precise information by encoding a single probability distribution over the variables of interest.

Real-world applications often deal with imprecise information. For example, the exact probability of an event may be
hard to assess
and thus it may be given by a probability interval, i.e., lower and upper probability bounds. Similarly, the precise dependency between two events may not be known and thus it may be described by an interval for a conditional probability. In general, there may be multiple sources of imprecise information: e.g., a Bayesian network learned from data as well as knowledge from human experts in the form of logic formulas annotated with confidence scores or probability intervals. Aggregating multiple sources of imprecise information in an effective manner is therefore critical for inference tasks.

Over the past decades, numerous formalisms have been proposed to represent imprecise information \cite{fuzzy,ds,nilsson,fagin1990logic,heinsohn1994probabilistic,jaeger1994probabilistic,andersen1994bayesian,hooker1999,cozman2000credal,cano2002using,durig2005probabilistic,problog,cozman2008cralc,lukasiewicz2008expressive,lnn}.
Some of these formalisms do not use probability functions to represent uncertainty.

This paper builds on earlier proposals for probabilistic logic \cite{nilsson,fagin1990logic,nilsson2}, credal networks \cite{andersen1994bayesian,cozman2000credal,cano2002using} and their variants \cite{cozman2008cralc,cozman2009}.
A common feature of these formalisms is that imprecise information is typically expressed by probability upper/lower bounds, which are treated as constraints that must be satisfied by each probability distribution in the entailed set of distributions. In many practical real-world situations, these upper and lower probability values have intuitive frequentist interpretations and therefore can be naturally incorporated in a probabilistic logic program \cite{nilsson,cozman2008cralc}. 

We highlight two important yet contrasting features of the above formalisms. On the one hand,
some probabilistic logics
\cite{nilsson,fagin1990logic,nilsson2} impose few restrictions on the logic formulas and therefore are quite expressive. They however lack
independence declarations,
which leads to excessively wide intervals in the inference results. Consequently, they are less useful for decision making in real-world applications. On the other hand, credal networks and their variants inherit the Markov condition of Bayesian networks \cite{pearl88}. Yet these models either can only express the structure of a Bayesian network \cite{andersen1994bayesian,cozman2000credal,cano2002using} or require acyclicity and other strong restrictions on logic formulas \cite{cozman2008cralc,cozman2009}.

\textbf{Contribution:~}
In this paper, we present \emph{Logical Credal Networks} (LCN), a new probabilistic logic model designed to exploit the best of both worlds. On one hand, the model allows probability bounds and conditional probability bounds for arbitrary propositional and finite-domain first-order logic formulas without requiring acyclicity. On the other hand, we define a Markov condition that allows random variables to have cyclic dependencies and yet we show that our proposed Markov condition matches the Markov condition in Bayesian/credal networks for acyclic graphs. Subsequently, we describe exact and approximate inference algorithms to compute the posterior probability of a query formula. We evaluate the proposed model on a maximum a posteriori (MAP) inference task using benchmark problems derived from Mastermind games with uncertainty and a realistic credit card fraud detection application. Our experimental results are quite promising and show that the proposed method outperforms existing approaches. In particular, they highlight the ability of LCNs to aggregate multiple sources of imprecise information in an effective manner.

%% file: background.tex
\section{Background} \label{sec:bg}

\subsection{Bayesian and Credal Networks}
A \emph{Bayesian network} (BN) \cite{pearl88} is defined by a tuple $\langle \bX, \bD, \bP, G \rangle$, where $\bX=\{X_1, \ldots, X_n\}$ is a set of variables over multi-valued domains $\bD=\{D_1, \ldots, D_n\}$, $G$ is a directed acyclic graph (DAG) over $\bX$ as nodes, and $\bP = \{P_i\}$ where $P_i = P(X_i|pa(X_i))$ are \emph{conditional probability tables} (CPTs) associated with each variable $X_i$ and $pa(X_i)$ are the parents of $X_i$ in $G$. The \emph{Markov condition} in BNs states that each node is independent of its non-descendants given its parents. Consequently, the joint probability distribution $P(\bX)$ is $ \prod_{i=1}^{n} P(X_i|pa(X_i))$. 

Bayesian networks are typically formulated and learned using a combination of expert opinion and tabular data. Prior work has also studied other forms of knowledge to construct Bayesian networks, besides structural information and point estimates of conditional probabilities. For instance, there is a line of research on exploiting qualitative information to learn the parameters of a Bayesian network~\cite{wellman1990, druzdzel1995,wittig2000}.

Credal networks \cite{andersen1994bayesian,cozman2000credal,cano2002using} extend Bayesian networks to deal with imprecise probabilities. A credal set is a set of probability distributions. A \emph{credal network} (CN) is defined by a pair $(G, \mathbb{K})$, where $G$ is a DAG over discrete variables $\bX$ and $\mathbb{K}$ is a set of conditional credal sets $K(X_i|pa(X_i))$ each one associated with a variable $X_i \in \bX$ and its parents $pa(X_i)$ in $G$. We consider \emph{separately specified} credal networks where each variable $X_i$ and each configuration $\pi_{ik}$ of  $pa(X_i)$ has a conditional credal set $K(X_i|pa(X_i)=\pi_{ik})$ which is specified separately from all others.

The following Markov condition for credal networks implies that two variables $X$ and $Y$ are independent iff the vertices of the credal set $K(X,Y)$ factorize, i.e., each distribution $P(X,Y)$ that is a vertex of the set $K(X,Y)$ satisfies $P(X,Y)=P(X)\cdot P(Y)$ for all values of $X$ and $Y$ (and likewise for conditional independence). Similar to how Bayesian networks specify a joint probability distribution over variables $\bX$, credal networks can be used to specify a set of joint probability distributions over $\bX$. The largest extension of a credal network that complies with the Markov condition above is called the \emph{strong extension}
of the network: $\{\prod_{i=1}^{n} P(x_i|pa(X_i)) ~:~ P(X_i|pa(X_i)) \in K(X_i|pa(X_i)) \}$ \cite{cozman2000credal}.

A common representation of credal sets which we also assume in this paper are probability intervals.

\subsection{Propositional and First-Order Theories}
In propositional logic, propositions can take only two values $\{True, False\}$ and are denoted by lowercase letters such as $x, y,$ and $z$. Propositional literals (i.e., $x$ , $\neg x$ ) stand for $x$ being True or $x$ being False. In first-order logic (FOL), a \emph{term} is a variable, a constant, or a function applied to terms. An \emph{atom} (or atomic formula) is either a proposition or a predicate $p(t_1,\ldots,t_n)$ of arity $n$ where the $t_i$ are terms. A \emph{formula} (either propositional or FOL) is built out of atoms using universal and existential quantifiers (for FOL) and the usual logical connectives $\neg$, $\vee$, $\wedge$ and $\rightarrow$, respectively.

\subsection{Probabilistic Logics}
\label{sec-prob-logic}
Syntactically, a probabilistic logic program is a set of logic formulas (either propositional or FOL), each one of them being annotated with a probability value \cite{nilsson}. Let $(q, \pi_q)$ be a pair such that $q$ is a formula and $\pi_q \in [0,1]$ is the associated probability value. The semantics is the set of probability distributions over all interpretations such that $P(q) = \pi_q$\footnote{Throughout this paper, we use $P\left( q \right)$ as shorthand notation for $P\left( q \text{ is True} \right)$ and $P\left( q \mid r \right)$ for $P\left( q\text{ is True} \mid r\text{ is True} \right)$.}. 
Point values $\pi_q$ can be easily replaced by intervals.
Each inference is a pair of optimization problems to compute the upper and lower bounds of the probability of interest.
The logic in \citet{fagin1990logic} is more general, and for example can represent bounds of conditional probabilities which are important in real-world tasks \cite{nilsson2}.

A major weakness of these probabilistic logics is the lack of a Markov condition: there are no independence relations that are implied in a probabilistic logic program. Consider, for example, the following simple logic program:
\begin{align}
  0.3 \leq& P\left( x \right) \leq 0.7 \label{eq:eg1}\\
  0.3 \leq& P\left( y \right) \leq 0.7 \label{eq:eg2}
\end{align}
where $x,y$ are atomic formulas. Following \citet{nilsson}, computing the bounds on $P(x\oplus y)$\footnote{Symbol $\oplus$ stands for XOR.} results in the interval $\left[0,1\right]$.
Indeed, there exists a joint distribution over $x,y$ such that (\ref{eq:eg1}) and (\ref{eq:eg2}) are satisfied and that $x \oplus y$ is always false, and there exists another such that $x \oplus y$ is always true.
The inference result of $\left[0,1\right]$, however, is not informative for most purposes and often is not the intention when one writes down (\ref{eq:eg1})(\ref{eq:eg2}) for an application.
This example illustrates that due to the lack of a Markov condition, arbitrary dependence between variables is considered possible, which results in excessively wide intervals in inference results.
More recent works on description logic \cite{heinsohn1994probabilistic,jaeger1994probabilistic,lukasiewicz2008expressive} share the same weakness.
A more practical approach in this case is to assume that $x$ and $y$ are independent of each other unless there is information saying otherwise. With independence assumption, computing bounds for $P(x \oplus y)$ results in the interval $\left[0.42,0.58\right]$.

A straightforward way to allow for a Markov condition in a probabilistic logic program is to treat it as a credal network with probability intervals. Consider the following probabilistic logic program / credal network:
\begin{align}
  0.3 \leq& P\left( x \right) \leq 0.7 \label{eq:eg2_1}\\
  0.1 \leq& P\left( y \mid x \right) \leq 0.2 \label{eq:eg2_2}\\
  0.6 \leq& P\left( y \mid \neg x \right) \leq 0.7 \label{eq:eg2_3}\\
  0.3 \leq& P\left( z \mid y \right) \leq 0.4 \label{eq:eg2_4}\\
  0.8 \leq& P\left( z \mid \neg y \right) \leq 0.9 \label{eq:eg2_5}
\end{align}
where $x,y,z$ are atomic formulas (i.e., binary variables). The Markov condition in this credal network is that $z$ is independent of $x$ given $y$, in mathematical terms:
\begin{align}
  P\left( z \mid y \right)      =& P\left( z \mid y      \land x \right) \label{eq:eg2_6}\\
  P\left( z \mid \neg y \right) =& P\left( z \mid \neg y \land x \right) \label{eq:eg2_7}
\end{align}
which is the network's strong extension \cite{cozman2000credal}.

Clearly, the credal network above represents the set of probability distributions over all interpretations such that constraints (\ref{eq:eg2_1}--\ref{eq:eg2_7}) are satisfied. However, this representation comes with several restrictions: (1) the only non-atomic logic formulas allowed are AND over atomic formulas and negation of atomic formulas; (2) there must not be cyclic dependencies among atomic formulas; (3) an atomic formula must be specified by either a marginal probability interval or by a set of conditional probability intervals, and not both; (4) the conditions in the conditional probabilities must enumerate all possible interpretations of the parent variables -- we will refer to the last two requirements as the \emph{unique-assessment assumption}. In practice, there is often knowledge that cannot be expressed by a simple AND, and there are often multiple sources of information that, when aggregated, break the acyclicity or unique-assessment requirements. More recently, \citet{cozman2008cralc,cozman2009} relax some of the restrictions above and allow for example to specify a conditional probability interval for $P\left( q \mid r \right)$, where $r$ can be an arbitrary logic formula but $q$ is an atom.

%% file: lcn.tex
\section{Logical Credal Networks}
\label{sec-lcn}

In this section, we introduce a new probabilistic logic called a \emph{Logical Credal Network} (LCN), that is designed to have the best of both worlds, namely as few restrictions as possible on logic formulas when specifying probability bounds and a set of implied independence relations that are similar to the Markov condition in Bayesian and credal networks.

\subsection{Syntax}

Syntactically, an LCN is specified by a set of \emph{probability-assessment sentences} in one of the following two forms:
\begin{align}
  l_q \leq& P\left( q \right) \leq u_q \label{eq:syn1}\\
  l_{q \mid r} \leq& P\left( q \mid r \right) \leq u_{q \mid r} \label{eq:syn2}
\end{align}
where $q$ and $r$ can be arbitrary propositional and finite-domain first-order logic formulas and $0 \leq l_q \leq u_q \leq 1$, $0 \leq l_{q \mid r} \leq u_{q \mid r} \leq 1$. Each sentence is further associated with a Boolean label $\tau\in\left\{ True, False \right\}$, which indicates whether formula $q$ implies graphical dependence and will be explained in the next section.

\subsection{Semantics}

An LCN represents the set of all its models. A model\footnote{The semantics is not model-theoretic because there exist implied constraints that are jointly derived from multiple sentences.} of an LCN is a probability distribution over all interpretations such that it satisfies a set of constraints given explicitly by (\ref{eq:syn1})--(\ref{eq:syn2}) and a set of independence constraints which are implied by the LCN. The latter are similar to the independence relations implied by a Markov condition in graphical models.
It is important that the independence constraints are implied rather than explicitly stated, because requiring a user to explicitly specify independence constraints would be tedious and potentially error-prone in a real-world application.

Therefore, the critical aspect of LCN semantics is how to define the implied independence constraints. In contrast to previous work \cite{andersen1994bayesian,cozman2000credal,cozman2009}, we propose a generalized Markov condition that accommodates the LCN's much more relaxed requirements on logic formulas, including cyclic dependencies. In particular, our definition is backward compatible and matches the Markov condition in Bayesian networks when (\ref{eq:syn1})--(\ref{eq:syn2}) happen to specify the marginal and conditional probabilities of a Bayesian network.

\begin{figure}[t!]
  \centering
  \includegraphics[width=\columnwidth]{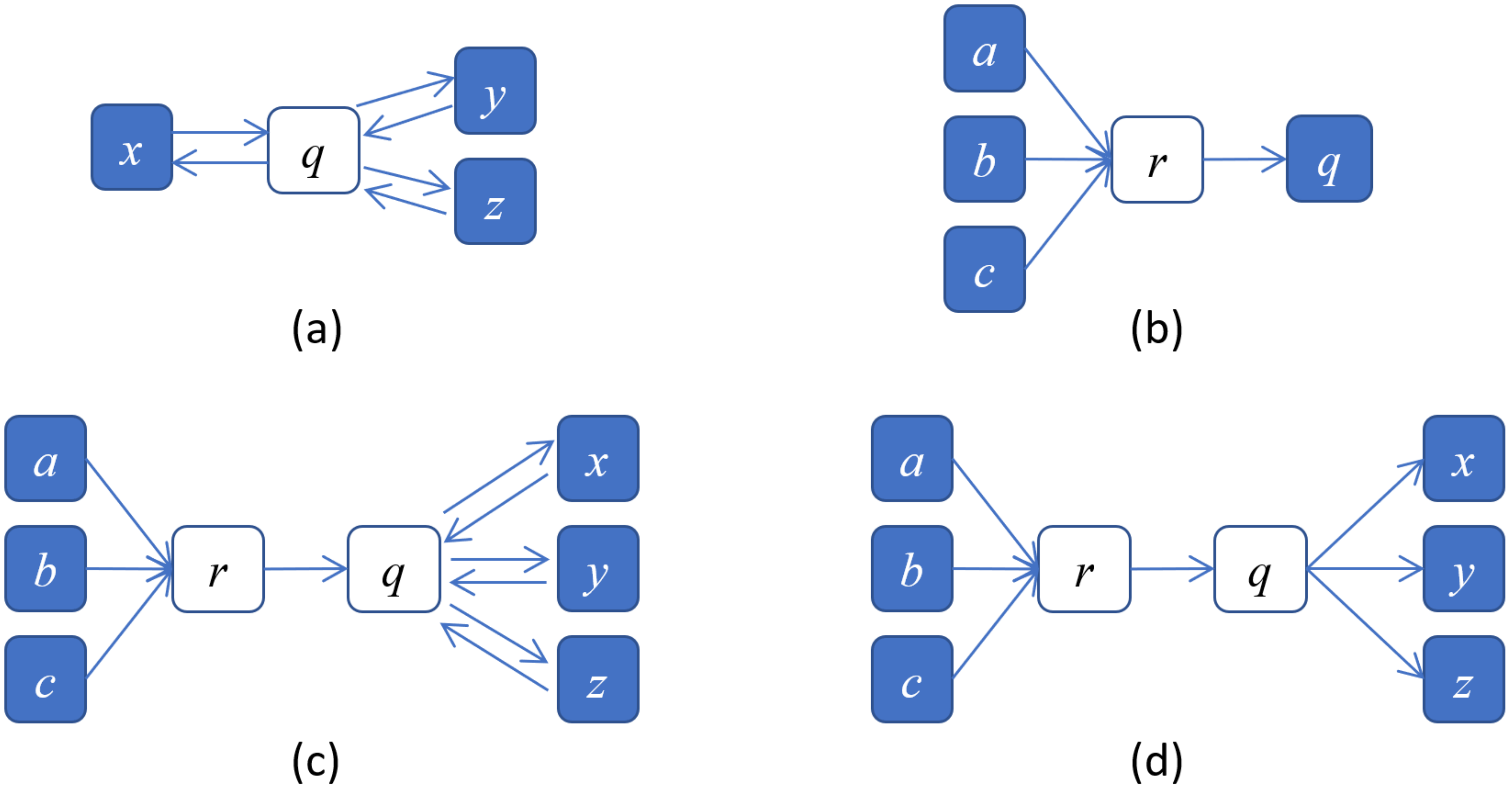}
  \caption{Stamps of sentences in LCNs: (a) A sentence (10) with $\tau = True$; (b) A sentence (11) where $q$ is an atomic formula; (c) A sentence (11) where $q$ is a non-atomic formula and $\tau = True$; (d) A sentence (11) where $q$ is a non-atomic formula and $\tau = False$. Atomic nodes are shaded.}
  \label{fig:stamp}
\end{figure}

We begin by defining the \emph{dependency graph} of an LCN. Given an LCN $\cM$, its dependency graph $G$ contains a node for each atomic and non-atomic formula in $\cM$. A sentence in $\cM$ induces a set of directed edges in $G$ which we call a \emph{stamp}. Assuming $\tau=True$, then for each sentence (\ref{eq:syn1}) such that $q$ has $x_1, \ldots, x_n$ atomic formulas, its stamp contains directed edges from $q$ to each $x_i$, and from each $x_i$ to $q$, respectively. Similarly, for each sentence (\ref{eq:syn2}) such that $x_1, \ldots, x_n$ (resp. $a_1, \ldots, a_m$) are the atomic formulas in $q$ (resp. $r$), its stamp contains a directed edge from $r$ to $q$, a set of directed edges from each $a_j$ to $r$, as well as directed edges from $q$ to each $x_i$ and from each $x_i$ to $q$, respectively. Setting $\tau$ to $False$ indicates that the sentences (\ref{eq:syn1}) or (\ref{eq:syn2}) do not imply graphical dependency among atomic formulas in $q$. In this case, the stamp of (\ref{eq:syn1}) would be empty, while the stamp of (\ref{eq:syn2}) would not involve directed edges from $x_i$ to $q$. Figure \ref{fig:stamp} illustrates stamps of the two types of sentences in LCNs, for $\tau=True$ and $\tau=False$, respectively.

\begin{figure*}[t!]
  \centering
  \includegraphics[width=1.5\columnwidth]{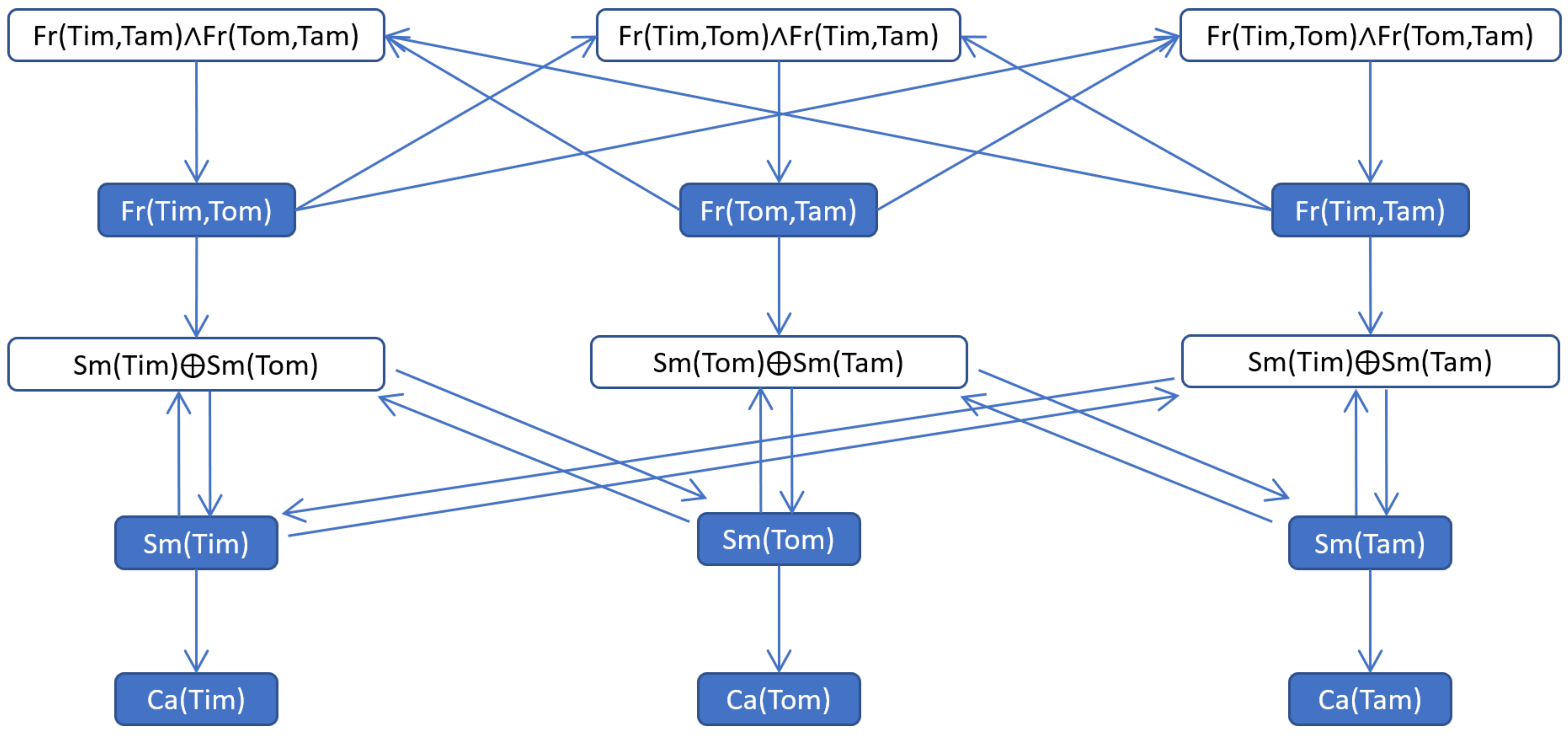}
  \caption{An example dependency graph. Atomic nodes are shaded.}
  \label{fig:smoke}
\end{figure*}

The intuition behind the stamps is the need to capture two types of dependencies. For a sentence (\ref{eq:syn1}) or (\ref{eq:syn2}) with $\tau=True$, the dependency among atomic formulas in $q$ is similar to a clique in Markov random fields (see the bi-directional edges between $q$ and its atomic formulas in Figures~\ref{fig:stamp}(a)(c)). For a sentence (\ref{eq:syn2}), the dependency between $q$ and $r$ is similar to the dependencies in Bayesian networks (see the one-directional connections in Figures~\ref{fig:stamp}(b)(c)(d))

\begin{example}\label{eg:smoke}
Consider the following LCN derived from the Smokers and Friends example of \citet{mln}. We abbreviate the predicates $Friends(\cdot,\cdot)$, $Smokes(\cdot)$ and $Cancer(\cdot)$ by $Fr(\cdot,\cdot)$, $Sm(\cdot)$ and $Ca(\cdot)$, respectively. Predicate $Fr(\cdot,\cdot)$ is symmetric.
\begin{align}
  0.5 \leq& P\left( \text{Fr}\left( \alpha,\gamma \right) \mid \text{Fr}\left( \alpha,\beta \right) \land \text{Fr}\left( \beta,\gamma \right)  \right) \leq 1,\forall \alpha \forall \beta \forall \gamma \label{eq:smf1}\\
  0 \leq& P\left( \text{Sm}\left( \alpha \right) \oplus \text{Sm}\left( \beta \right) \mid \text{Fr}\left( \alpha,\beta \right) \right) \leq 0.2, \forall \alpha \forall \beta \label{eq:smf2} \\
  0.03 \leq& P\left( \text{Ca}\left( \alpha \right) \mid \text{Sm}\left( \alpha \right) \right) \leq 0.04, \forall \alpha \label{eq:smf3} \\
  0 \leq& P\left( \text{Ca}\left( \alpha \right) \mid \neg \text{Sm}\left( \alpha \right) \right) \leq 0.01, \forall \alpha \label{eq:smf4}
\end{align}
Here, (\ref{eq:smf1}) states that friends of friends are likely friends; (\ref{eq:smf2}) states that, if two people are friends, they likely either both smoke or neither does; (\ref{eq:smf3}) and (\ref{eq:smf4}) state that smoking likely causes cancer.
$\tau=True$ for all sentences.
Figure~\ref{fig:smoke} illustrates the dependency graph of the LCN grounded on a domain of three people (as before, symbol $\oplus$ is XOR).
\end{example}

With the dependency graph, we are ready to define the generalized Markov condition for LCNs.

\begin{definition}
  The parents of an atomic formula $x$, denoted by $\text{parents}\left( x \right)$, are the set of atomic formulas such that there exists a directed path in the dependency graph from each of them to $x$ in which all intermediate nodes are non-atomic.
\end{definition}

\begin{definition}
  The descendants of an atomic formula $x$, denoted by $\text{descendants}\left( x \right)$, are the set of atomic formulas such that there exists a directed path in the dependency graph from $x$ to each of them in which no intermediate node is in $\text{parents}\left( x \right)$.
\end{definition}

Let $\text{ndnp}\left( x \right)$ denote the non-descendant non-parent variables of $x$: $\text{ndnp}\left( x \right) \triangleq \left\{\text{atomic formulas}\right\} \setminus \text{parents}\left( x \right) \setminus \text{descendants}\left( x \right) \setminus \left\{ x \right\}$.

\begin{definition}[Markov condition]
In a model of an LCN, every atomic formula $x$ is conditionally independent of $\text{ndnp}\left( x \right)$ given $\text{parents}\left( x \right)$.
\end{definition}

In Example~\ref{eg:smoke} and Figure~\ref{fig:smoke}, the Markov condition includes that $\text{Sm}\left( \text{Tim} \right)$ is conditionally independent of $\text{Fr}\left( \text{Tom},\text{Tam} \right)$, $\text{Ca}\left( \text{Tom} \right)$ and $\text{Ca}\left( \text{Tam} \right)$ given $\text{Fr}\left( \text{Tim},\text{Tom} \right)$, $\text{Fr}\left( \text{Tim},\text{Tam} \right)$, $\text{Sm}\left( \text{Tom} \right)$ and $\text{Sm}\left( \text{Tam} \right)$, and that $\text{Ca}\left( \text{Tim} \right)$ is conditionally independent of all other variables given $\text{Sm}\left( \text{Tim} \right)$.

\paragraph{Remark}
Consider the special case of using LCNs to represent Bayesian networks.
The upper and lower bounds in a LCN sentence (\ref{eq:syn1}) or (\ref{eq:syn2}) are equal; formulas $q$ in (\ref{eq:syn1})(\ref{eq:syn2}) are atomic; and the LCN sentences specify the marginal and conditional probabilities in a Bayesian network.
The LCN dependency graph is constructed by stamps only in the form of Figure~\ref{fig:stamp}(b).
Consequently, the parents of an atomic node $x$ are the same as the parents of $x$ in the Bayesian network, and the descendants of an atomic formula $x$ are the same as the descendants of $x$ in the Bayesian network.
Therefore, the Markov condition of the LCN is identical to that of the Bayesian network.
The LCN has only one model, which is the same probability distribution as the Bayesian network.

\subsection{Inference}

As previous probabilistic logics discussed in Section~\ref{sec-prob-logic}, inference in an LCN means computing upper and lower bounds on a probability of interest. This entails solving a pair of optimization problems comprising the constraints stated explicitly by the LCN sentences (\ref{eq:syn1})(\ref{eq:syn2}) and those derived from the generalized Markov condition. Specifically, (\ref{eq:syn1})(\ref{eq:syn2}) are linear constraints while constraints from the Markov condition are quadratic. The objective function can be in the form of either $P\left( q \right)$, where $q$ is a logic formula, or $P\left( q \mid e \right)$, where $q,e$ are logic formulas and $e$ represents evidence. In some scenarios we may be interested in the model with the maximum entropy \cite{cheeseman1983} and the objective function can be a measure of entropy. The appendix includes examples of these optimization problems.

\paragraph{Exact Inference}
For exact inference, we use the IPOPT solver \cite{ipopt} to solve the optimization problems.
The experiments in Section~\ref{sec:results} have been conducted with this approach.
Since the optimization problems are over the space of joint distributions, the complexity is exponential with respect to the number of atomic formulas, same as in prior works \cite{hooker1999}.
The experimental results suggest that, although with problem size limitations, the exact approach is able to perform inference for meaningful tasks such as solving Mastermind puzzles and credit card fraud detection.

For large scale problems, approximate inference algorithms are needed.
In order to handle imprecise probabilities, the classical belief propagation algorithms \cite{pearl82,weiss2000} for probabilistic graphical models have been extended to propagate intervals.
Specifically, the 2U algorithm \cite{2u} is exact on credal networks that are polytrees; the L2U and IPE algorithms \cite{l2u} are built on top of 2U for approximate inference on credal networks with loops; \citet{antonucci2010} generalize L2U to beyond binary variables.
\citet{cozman2009} suggested that L2U is the inference algorithm of choice for the general case of its semantics.

\paragraph{Incompatibility with Belief Propagation}
A closer examination however reveals that L2U, or any other method based on sum-product message passing, is incompatible with LCNs or work by \citet{cozman2008cralc,cozman2009} when the unique-assessment assumption is broken.
Consider the following example:
\begin{align}
  0.2 &\leq P\left(a\right) \leq 0.3 \label{eq:eg3_1}\\
  0.6 &\leq P\left(b\mid a\right) \leq 0.7 \label{eq:eg3_2}\\
  0.1 &\leq P\left(b\mid \neg a \right) \leq 0.2 \label{eq:eg3_3}\\
  0.3 &\leq P\left(b\right) \leq 0.4 \label{eq:eg3_4}
\end{align}
As discussed in Section~\ref{sec:bg}, the unique-assessment assumption is a property of credal/Bayesian networks.
Here, it allows (\ref{eq:eg3_2})(\ref{eq:eg3_3}) or (\ref{eq:eg3_4}), but not both.
Therefore, this example is not a credal network but is legitimate as an LCN or by the semantics of \citet{cozman2008cralc,cozman2009}.
Under both semantics, if we query $P\left(b\right)$, the correct answer is [0.3,0.35].
However, 2U or L2U gives an incorrect answer of [0.1,0.26], even though the dependency graph is a polytree.

The inconsistency stems from the fact that sum-product message passing is fundamentally a Markov random field solver and 2U/L2U treats probability values as factor potentials.
When the unique-assessment assumption is upheld, as in credal/Bayesian networks, products of factor potentials coincide with probabilities and therefore L2U works for credal networks.
As soon as the unique-assessment assumption is broken, the solvers lose correctness regardless of network topology.
Hence 2U/L2U/IPE are not fit for LCNs and not fit for the semantics of \cite{cozman2008cralc,cozman2009}.
In real-world applications, the unique-assessment assumption is often broken when multiple sources of information are aggregated, hence there is a need for approximate inference algorithms that can handle such scenarios.

\paragraph{Modified Belief Propagation}
We propose a modified belief propagation algorithm.
The high-level flow is identical to classical belief propagation.
We build a factor graph \cite{frey2003} with variable nodes and factor nodes, and iteratively update messages between variables and factors until convergence.
The difference lies in how messages are computed, which are not by sum and product but by tightening bounds at variable nodes and solving a local constraint program at factor nodes.
The discussion is limited to binary variables.
For multi-valued variables, messages will have more complex structures \cite{antonucci2010}.

An LCN factor graph is a bipartite graph with variable nodes, which represent atomic formulas, and factor nodes, each of which represents one or more sentences in the LCN.
Sentences that involve the same set of atomic formulas are grouped into one factor.
For example, an LCN by (\ref{eq:eg2_1}--\ref{eq:eg2_5}) has three factors: (\ref{eq:eg2_1}), (\ref{eq:eg2_2})(\ref{eq:eg2_3}) and (\ref{eq:eg2_4})(\ref{eq:eg2_5}).
Let $v$ denote a variable node.
Let $f$ denote a factor node.
Let $N\left(\cdot\right)$ denote the neighbors of a node.
A message is an interval $\left[l,u\right]$ where $0\leq l \leq u \leq 1$.
Let $\left[l_{v \rightarrow f},u_{v \rightarrow f}\right]$ denote the message from $v$ to $f$ and $\left[l_{f \rightarrow v},u_{f \rightarrow v}\right]$ from $f$ to $v$.

If a variable node has degree one, it sends a message of $\left[0,1\right]$ to its only factor neighbor.
If the degree of $v$ is more than one, it sends the following message to neighbor $f$:
\begin{align}
  l_{v \rightarrow f} &= \max_{f^\prime\in N\left(v\right)\setminus\{f\}} l_{f^\prime \rightarrow v} \\
  u_{v \rightarrow f} &= \min_{f^\prime\in N\left(v\right)\setminus\{f\}} u_{f^\prime \rightarrow v}
\end{align}

A factor node $f$ computes its message to neighbor $v$ by solving a local constraint program, which is composed of:
\begin{itemize}
\item
  Sentences of factor $f$;
\item
  $l_{v^\prime \rightarrow f} \leq P\left( v^\prime \right) \leq u_{v^\prime \rightarrow f},\forall v^\prime\in N\left(f\right)\setminus\{v\}$;
\item
  Quadratic constraints that the variables in $N\left(f\right)\setminus\{v\}$ are independent of each other.
\end{itemize}
The objective function is $P\left( v \right)$, and the message $l_{f \rightarrow v}$ and $u_{f \rightarrow v}$ are the results of minimizing and maximizing the objective with the local constraint program.

All messages are initialized to be $\left[0,1\right]$.
Convergence is guaranteed because the lower bound of each message monotonically increases while the upper bound of each message monotonically decreases.
The role of variable nodes is to tighten the bounds, which is fundamentally different from that in previous belief propagation methods.
More details and illustrating examples are in the appendix.

\paragraph{Comparison with Other Algorithms}
The bound tightening operation at variable nodes is similar to the inference algorithm in \citet{lnn}.
The message computation at factor nodes bears some similarity to 2U; the independence assumption in local constraint programs is a mechanism to approximate the Markov condition, and the same approach is used in classical belief propagation.
There is also a relation to the IPE algorithm \cite{l2u}: IPE cuts out a number of polytree subgraphs, solves each subgraph, and then chooses the tightest bounds from the subgraphs; our proposed algorithm implicitly enumerates an exponential number of subgraphs and chooses one subgraph for each $l$ or $u$ that computes the tightest bound.
The algorithm guarantees exact results on polytree credal/Bayesian networks, with and without additional marginal probability sentences that break the unique-assessment assumption.
In general, however, there is no guarantee on correctness; this is the same as classical belief propagation.

%% file: results.tex
\section{Experiments}
\label{sec:results}

We evaluate the proposed approach on the following variants of MAP
inference for LCNs. Given an LCN and a subset of query (or MAP)
variables, the task is to find the assignment to the query variables
such that its posterior marginal probability interval has the largest
upper bound (maximax) or, alternatively, the largest lower bound
(maximin).

\subsection{Mastermind Puzzles with Uncertainty}

Mastermind is a classic two-player code breaking game, made popular in computer science by~\citet{knuth1976}.
We consider Mastermind games with uncertainty which differ from the
classic game in that the code-maker lies randomly. If given
the probabilities of lying, a game is completely specified by a
Bayesian network. A large number of puzzles, i.e., game boards, are
generated by sampling from such Bayesian networks. Given imprecise
information of the underlying Bayesian network of each puzzle, the
task is to guess the hidden code. The ground truth is the MAP hidden
code(s) given the board as observations.
Details on puzzle generation and an example puzzle are in
the appendix.

Let $K$ be the number of rounds in a puzzle and let $l_i$ be a boolean
variable indicating whether the code-maker lied in the $i^\mathrm{th}$
round. Given a puzzle, the imprecise information available to the
inference algorithms is, for example:
\begin{align}
0.3   \leq & P\left( l_i \right) \leq 0.7 \,,\, \forall 1\leq i \leq K \label{eq:lieinterval}\\
0.245 \leq & P\left( l_1 \land l_2 \right) \leq 0.360 \label{eq:kn1}\\
0.795 \leq & P\left( l_2 \lor  l_3 \right) \leq 0.903 \label{eq:kn2}\\
0.207 \leq & P\left( l_3 \land l_4 \right) \leq 0.273 \label{eq:kn3}\\
& \quad\quad\cdots \nonumber
\end{align}
Equation (\ref{eq:lieinterval}) is a typical credal network
specification and is used by all inference algorithms, while only a
subset of the algorithms are able to utilize the remaining equations.
We use ten random seeds to generate ten sets of puzzles (each set
having 729 puzzles on average) and report the mean and standard
deviation of the accuracy (Table \ref{tab:mm}).

We consider the following competing methods:

\begin{table}[t!]
  \centering
  \begin{tabular}{l|c}
    \toprule
Method  & Accuracy \\
\midrule
Bayesian\_midpoint & 65.7\% $\pm$ 2.2\% \\
Credal\_maximax & 60.3\% $\pm$ 2.3\% \\
Credal\_maximin & 64.4\% $\pm$ 2.2\% \\
ProbLog\_midpoint & 65.7\% $\pm$ 2.2\% \\
cProbLog\_midpoint & 65.3\% $\pm$ 2.1\% \\
MLN\_midpoint & 65.3\% $\pm$ 2.1\% \\
Nilsson\_maximax & intractable \\
Nilsson\_maximin & 0\% \\
LCN\_maximax & 69.2\% $\pm$ 1.8\% \\
LCN\_maximin & 71.1\% $\pm$ 1.9\% \\
LCN\_maxent  & {\bf 72.5}\% $\pm$ 1.7\% \\
    \bottomrule
  \end{tabular}
  \caption{Accuracy on Mastermind puzzles.}
\label{tab:mm}
\end{table}

\textbf{Bayesian\_midpoint} simply assumes that $P\left( l_i \right)$
is equal to the midpoint of the interval specified in
(\ref{eq:lieinterval}) and performs MAP inference on the resulting
Bayesian network. It does not utilize logic formulas like
(\ref{eq:kn1})--(\ref{eq:kn3}).

\textbf{Credal\_maximax} and \textbf{Credal\_maximin} assume a credal
network specified by (\ref{eq:lieinterval}) and do not utilize logic
formulas like (\ref{eq:kn1})--(\ref{eq:kn3}). They compute the maximax
and the maximin MAP assignments respectively.

\textbf{ProbLog\_midpoint} and \textbf{cProbLog\_midpoint} are ProbLog
\cite{problog} and cProbLog \cite{cproblog} that use the midpoints of
the intervals because they do not allow probability intervals.
ProbLog is unable to represent formulas like (\ref{eq:kn1})--(\ref{eq:kn3}) and hence is the same as Bayesian\_midpoint.
cProbLog is able to represent (\ref{eq:kn1})--(\ref{eq:kn3}) as soft constraints.% and handle the resulting conflicts.

\textbf{MLN\_midpoint} is the Markov Logic Network (MLN) \cite{mln}
that uses all logic formulas as well as the
midpoint $p_\mathrm{mid}$ of each interval. Following
\citet{cproblog}, each formula is associated with weight
$w=\log\left( p_\mathrm{mid} / \left( 1 - p_\mathrm{mid} \right)
\right)$, and with these weights MLN\_midpoint is equivalent to cProbLog\_midpoint.
Because each puzzle is sampled from a different underlying
Bayesian network, it is impossible to train the MLN weights.

\textbf{Nilsson\_maximax} and \textbf{Nilsson\_maximin} are the
original probabilistic logic formulations from \citet{nilsson,nilsson2}
that compute the maximax and maximin MAP assignments  respectively.
Although Nilsson's method is able to utilize all
the logic formulas and their bounds, it does not allow modeling $l_i$
variables as being independent of each other. Given a puzzle, each
possible hidden code $c$ corresponds to a truth assignment to the
$l_i$ variables; without independence relations, there always exists a
joint probability distribution of the $l_i$ variables such that the
particular truth assignment has zero probability. Consequently, the
lower bound of the posterior probability given a puzzle is zero for
any $c$, and therefore the maximin criterion has no basis to make
decisions upon.  Intuitively, a similar effect should also impair the
accuracy of Nilsson\_maximax. However, we are unable to verify this
empirically: computing the upper bound of
the posterior by solving Nilsson's constraint program is
computationally prohibitive due to the complexity of the objective
function and the size of the program.

\textbf{LCN\_maximax} and \textbf{LCN\_maximin} are the proposed
methods computing the maximax and maximin MAP
assignments respectively.  In this case, sentences like (\ref{eq:kn1})--(\ref{eq:kn3})
are annotated with $\tau = False$ so that they do not imply dependency
among the $l_i$ variables.

\textbf{LCN\_maxent} is a variant of the proposed method that, given
all the constraints, computes a joint probability distribution of the
$l_i$ variables with the greatest entropy and assumes that it is the
underlying distribution.

Table \ref{tab:mm} demonstrates clearly that our proposed
approach outperforms all its competitors. Furthermore, it is the only method able to properly aggregate multiple
sources of knowledge specified as logic formulas with probability
bounds. We also notice that MLN\_midpoint and cProbLog\_midpoint are
slightly worse than Bayesian\_midpoint and ProbLog\_midpoint. cProbLog/MLN do not gain accuracy by exploiting additional
knowledge. This is because they treat each logic formula as a
statistically independent factor and the probabilities as factor
potentials; in other words, the probability midpoints are not treated
as probabilities in the joint distribution. This is a major weakness
of cProbLog and MLN, in addition to their inability to handle
intervals.  In contrast, the proposed LCN aggregates knowledge as
multiple sources of constraints and treats the given bounds as actual
probability bounds without making any unwarranted independence
assumptions among different pieces of knowledge.

\subsection{Credit Card Fraud Detection}
We consider a realistic credit card fraud detection task based on the
UCSD-FICO Data Mining Contest dataset \cite{ccdata} which contains
100,000 transactions over a period of 98 days out of which 2,654 are
fraudulent. Each transaction is characterized by transaction amount,
transaction time, state, hashed zip code, hashed email address
together with eleven other anonymized features.

We use the hashed email addresses as account IDs and split the data
into two parts.  The first contains 55,750 accounts, each with
a single transaction.  The second contains 14,374 accounts that
have multiple transactions, and the total number of transactions is
44,250. The first and second subsets form training and test data respectively.  In addition to learning from training data,
we provide additional knowledge regarding fraudulent transactions and account history through the following three logic rules
from \citet{li2020temporal}:
\begin{align}
\texttt{\scriptsize Is-Fraud}\left(t\right) \leftarrow & \texttt{\scriptsize Has-FraudHistory}\left(t'\right) \land \texttt{\scriptsize Before}\left(t',t\right) \nonumber\\
\texttt{\scriptsize Is-Fraud}\left(t\right) \leftarrow & \texttt{\scriptsize Has-ZeroAmountHistory}\left(t'\right) \land \texttt{\scriptsize Before}\left(t',t\right) \nonumber\\
\texttt{\scriptsize Is-Fraud}\left(t\right) \leftarrow & \texttt{\scriptsize Has-MultiZip}\left(t'\right) \land \texttt{\scriptsize Before}\left(t',t\right) \nonumber
\end{align}

We create ten randomized tasks by sampling half of
the training data and half of the test data for each.  Note that we
sample accounts rather than transactions; in other words, transactions
from the same account are either all included in a test set or all
excluded from it. Subsequently, we emulate expert knowledge as
follows.  For each of the three logic rules and each of the ten test
sets, we measure the conditional probability that the consequent is
true given that the antecedent is true. We take the min and max over
the test sets and obtain the following probability intervals for the
three rules: $\left[0.65,0.74\right]$,
$\left[0.31,0.66\right]$ and $\left[0.44,72\right]$,
respectively.

Prediction is based on a binary posterior distribution and consequently
the maximax and maximin decision criteria give the same results.
We consider the following methods.

\begin{table}[t!]
\centering
\begin{tabular}{l|c}
\toprule
Method  & F1 score \\
\midrule
Naive\_Bayes       & 0.408 $\pm$ 0.110  \\
Bayesian\_midpoint & 0.090  $\pm$ 0.015 \\
Credal             & 0.089 $\pm$ 0.015 \\
ProbLog\_midpoint  & 0.599 $\pm$ 0.048 \\
MLN\_midpoint      & 0.472 $\pm$ 0.094 \\
Nilsson            & intractable \\
LCN                & {\bf 0.630} $\pm$ 0.046 \\
\bottomrule
\end{tabular}
\caption{F1 scores on credit card fraud detection.}
\label{tab:cc}
\end{table}

\textbf{Naive\_Bayes} uses only the training data and does not use
expert knowledge.

\textbf{Bayesian\_midpoint} uses a Bayesian network that expands
Naive\_Bayes by adding the antecedents in the three logic rules as
three parent nodes of the Is-Fraud node. The midpoints of the three
intervals are used; they are not enough to fully specify the
dependencies between the three new nodes and the Is-Fraud node, and we
use the noisy-OR model to address this issue.  The prior probability
of the three new nodes is 0.5 which does not change the inference
result.

\textbf{Credal} uses a credal network with the same structure as
Bayesian\_midpoint but with the three intervals.  The same noisy-OR
model is applied and the prior on the three new nodes is the
probability interval $\left[0,1\right]$.  

\textbf{ProbLog\_midpoint} contains the Naive\_Bayes model expressed
in ProbLog and uses the three logic rules annotated with the midpoint
of the corresponding probability
interval.

\textbf{MLN\_midpoint} contains the Naive\_Bayes model expressed as an
MLN and adds the three logic rules as follows. Let $p_\mathrm{mid}$ be
the midpoint of one of the three intervals. We add a factor of
$\texttt{antecedent}\land\texttt{consequent}$ with weight
$\log\left( p_\mathrm{mid}\right)$ and another factor of
$\texttt{antecedent}\land\neg\texttt{consequent}$ with weight
$\log\left(1 - p_\mathrm{mid} \right)$, respectively.
Because the training data contains no account history, it is impossible to train the MLN weights.

\textbf{Nilsson} \cite{nilsson,nilsson2} cannot represent independence
relations and is therefore computationally prohibitive.

\textbf{LCN} is the proposed method.

Table \ref{tab:cc} reports the mean and standard deviation of the F1
scores obtained on the test set of the credit card fraud dataset. We see again that
our proposed LCN method substantially outperforms its competitors. Bayesian\_midpoint and Credal perform quite poorly in
this case because of the unique-assessment requirement of Bayesian/credal
networks. Specifically, since the \texttt{Is-Fraud} node has three parents,
we are no longer allowed to specify $P\left(\text{Is-Fraud}\right)$.
In the Naive Bayes models, $P\left(\text{Is-Fraud}\right)$ values are
measured on the training data. This information is lost in
Bayesian\_midpoint and Credal models and consequently, false positive
predictions increase dramatically.

%% file: sup.tex
\appendix

\numberwithin{equation}{section}
\numberwithin{table}{section}
\numberwithin{figure}{section}
\numberwithin{algorithm}{section}

\section{Examples of Exact Inference}

Consider the following LCN:
\begin{align}
  0.6 &\leq P\left(a \land b \right) \leq 1 \label{eq:eg4_1}\\
  0 &\leq P\left(a \mid c \right) \leq 0.2  \label{eq:eg4_2}\\
  0 &\leq P\left(a \mid \neg c \right) \leq 0.8  \label{eq:eg4_3}\\
  0 &\leq P\left(b \mid d \right) \leq 0.7  \label{eq:eg4_4}\\
  0 &\leq P\left(b \mid \neg d \right) \leq 0.3  \label{eq:eg4_5}
\end{align}
where $\tau = True$ for the sentences.
The implied constraints by the Markov condition are: $c$ and $d$ are independent; $a$ is conditionally independent of $d$ given $b$ and $c$; $b$ is conditionally independent of $c$ given $a$ and $d$.

Define sixteen variables $p_{0,0,0,0},p_{0,0,0,1},\cdots,p_{1,1,1,1}$ to represent the probabilities of the sixteen interpretations, where the four bits in subscript represent the truth values of $a$, $b$, $c$ and $d$ respectively.
For example, $p_{0,0,0,0}$ is the probability that $a$, $b$, $c$ and $d$ are all false.

Consider a query on the marginal probability of $P\left(c\right)$.
We formulate two optimization problems:
\begin{align}
& p_{i,j,k,l} \geq 0 \quad , \quad \forall i,j,k,l \in \left\{ 0,1 \right\} \label{eq:eg4_6}\\
& \sum_{i,j,k,l \in \left\{ 0,1 \right\}} p_{i,j,k,l} = 1 \label{eq:eg4_7}\\
& \sum_{k,l \in \left\{ 0,1 \right\}} p_{1,1,k,l} \geq 0.6 \label{eq:eg4_8} \\
& \sum_{j,l \in \left\{ 0,1 \right\}} p_{1,j,1,l} \leq 0.2 \cdot \sum_{i,j,l \in \left\{ 0,1 \right\}} p_{i,j,1,l} \\
& \sum_{j,l \in \left\{ 0,1 \right\}} p_{1,j,0,l} \leq 0.8 \cdot \sum_{i,j,l \in \left\{ 0,1 \right\}} p_{i,j,0,l} \\
& \sum_{i,k \in \left\{ 0,1 \right\}} p_{i,1,k,1} \leq 0.7 \cdot \sum_{i,j,k \in \left\{ 0,1 \right\}} p_{i,j,k,1} \\
& \sum_{i,k \in \left\{ 0,1 \right\}} p_{i,1,k,0} \leq 0.3 \cdot \sum_{i,j,k \in \left\{ 0,1 \right\}} p_{i,j,k,0} \label{eq:eg4_12} \\
& \sum_{i,j,l \in \left\{ 0,1 \right\}} p_{i,j,1,l} \cdot \sum_{i,j,k \in \left\{ 0,1 \right\}} p_{i,j,k,1} = \sum_{i,j \in \left\{ 0,1 \right\}} p_{i,j,1,1} \label{eq:eg4_13}\\
& \sum_{l \in \left\{ 0,1 \right\}} p_{1,0,0,l} \cdot \sum_{i \in \left\{ 0,1 \right\}} p_{i,0,0,1} = p_{1,0,0,1} \cdot \sum_{i,l \in \left\{ 0,1 \right\}} p_{i,0,0,l} \\
& \sum_{l \in \left\{ 0,1 \right\}} p_{1,0,1,l} \cdot \sum_{i \in \left\{ 0,1 \right\}} p_{i,0,1,1} = p_{1,0,1,1} \cdot \sum_{i,l \in \left\{ 0,1 \right\}} p_{i,0,1,l} \\
& \sum_{l \in \left\{ 0,1 \right\}} p_{1,1,0,l} \cdot \sum_{i \in \left\{ 0,1 \right\}} p_{i,1,0,1} = p_{1,1,0,1} \cdot \sum_{i,l \in \left\{ 0,1 \right\}} p_{i,1,0,l} \\
& \sum_{l \in \left\{ 0,1 \right\}} p_{1,1,1,l} \cdot \sum_{i \in \left\{ 0,1 \right\}} p_{i,1,1,1} = p_{1,1,1,1} \cdot \sum_{i,l \in \left\{ 0,1 \right\}} p_{i,1,1,l}
\end{align}
\begin{align}
& \sum_{k \in \left\{ 0,1 \right\}} p_{0,1,k,0} \cdot \sum_{j \in \left\{ 0,1 \right\}} p_{0,j,1,0} = p_{0,1,1,0} \cdot \sum_{j,k \in \left\{ 0,1 \right\}} p_{0,j,k,0} \\
& \sum_{k \in \left\{ 0,1 \right\}} p_{0,1,k,1} \cdot \sum_{j \in \left\{ 0,1 \right\}} p_{0,j,1,1} = p_{0,1,1,1} \cdot \sum_{j,k \in \left\{ 0,1 \right\}} p_{0,j,k,1} \\
& \sum_{k \in \left\{ 0,1 \right\}} p_{1,1,k,0} \cdot \sum_{j \in \left\{ 0,1 \right\}} p_{1,j,1,0} = p_{1,1,1,0} \cdot \sum_{j,k \in \left\{ 0,1 \right\}} p_{1,j,k,0} \\
& \sum_{k \in \left\{ 0,1 \right\}} p_{1,1,k,1} \cdot \sum_{j \in \left\{ 0,1 \right\}} p_{1,j,1,1} = p_{1,1,1,1} \cdot \sum_{j,k \in \left\{ 0,1 \right\}} p_{1,j,k,1} \label{eq:eg4_21}\\
& \text{maximize/minimize } \sum_{i,j,l \in \left\{ 0,1 \right\}} p_{i,j,1,l} \label{eq:eg4_obj}
\end{align}
Constraints (\ref{eq:eg4_6})(\ref{eq:eg4_7}) ensure that the sixteen variables are a valid probability distribution.
(\ref{eq:eg4_8}--\ref{eq:eg4_12}) are explicit and linear constraints from the LCN sentences (\ref{eq:eg4_1}--\ref{eq:eg4_5}).
(\ref{eq:eg4_13}--\ref{eq:eg4_21}) are implicit and quadratic constraints from the Markov condition, and they have been reduced using techniques from \cite{andersen1994bayesian}.
By maximizing and minimizing the objective function (\ref{eq:eg4_obj}), we obtain the upper and lower bounds for $P\left(c\right)$, which are 0.33 and 0 respectively.
For another query on the posterior probability of $P\left( a \mid b \right)$, we replace (\ref{eq:eg4_obj}) with the following objective:
\begin{equation}
    \frac{\sum_{k,l \in \left\{ 0,1 \right\}} p_{1,1,k,l}}{\sum_{i,k,l \in \left\{ 0,1 \right\}} p_{i,1,k,l}}
\end{equation}
and the resulting interval is $\left[ 0.85,1\right]$.
In some scenarios we may be interested in the model with the maximum entropy \cite{cheeseman1983} and therefore minimize the following objective instead.
\begin{equation}
    \sum_{i,j,k,l \in \left\{ 0,1 \right\}} p_{i,j,k,l} \cdot \log p_{i,j,k,l}
\end{equation}

\section{Details of the Modified Belief Propagation Algorithm}

Let's use the example from Section~3.3, copied here:
\begin{align}
  0.2 &\leq P\left(a\right) \leq 0.3 \label{eq:eg3b_1}\\
  0.6 &\leq P\left(b\mid a\right) \leq 0.7 \label{eq:eg3b_2}\\
  0.1 &\leq P\left(b\mid \neg a \right) \leq 0.2 \label{eq:eg3b_3}\\
  0.3 &\leq P\left(b\right) \leq 0.4 \label{eq:eg3b_4}
\end{align}
If we query $P\left(b\right)$, the correct answer is [0.3,0.35] according to both our semantics and the semantics in \cite{cozman2009}.
However, both the 2U \cite{2u} and L2U \cite{l2u} algorithms compute an incorrect answer of [0.1,0.26].

The high-level flow of the new algorithm is identical to classical belief propagation.
We build a factor graph with variable nodes and factor nodes, and iteratively update messages from variables to factors and messages from factors to variables until convergence.
The factor graph is a bipartite graph with variable nodes, which represent atomic formulas, and factor nodes, each of which represents one or more sentences in the LCN.
Sentences that involve the same set of atomic formulas are grouped into one factor.
For example, an LCN by (\ref{eq:eg3b_1}--\ref{eq:eg3b_4}) has three factors: $f_1$ is (\ref{eq:eg3b_1}), $f_2$ is (\ref{eq:eg3b_2})(\ref{eq:eg3b_3}), and $f_3$ is (\ref{eq:eg3b_4}).

Let $v$ denote a variable node.
Let $f$ denote a factor node.
Let $N\left(\cdot\right)$ denote the neighbors of a node.
A message is an interval $\left[l,u\right]$ where $0\leq l \leq u \leq 1$.
Let $\left[l_{v \rightarrow f},u_{v \rightarrow f}\right]$ denote the message from $v$ to $f$ and $\left[l_{f \rightarrow v},u_{f \rightarrow v}\right]$ from $f$ to $v$.
If a variable node has degree one, it sends a message of $\left[0,1\right]$ to its only factor neighbor.
If the degree of $v$ is more than one, it sends the following message to neighbor $f$:
\begin{align}
  l_{v \rightarrow f} &= \max_{f^\prime\in N\left(v\right)\setminus\{f\}} l_{f^\prime \rightarrow v} \\
  u_{v \rightarrow f} &= \min_{f^\prime\in N\left(v\right)\setminus\{f\}} u_{f^\prime \rightarrow v}
\end{align}
A factor node $f$ computes its message to neighbor $v$ by solving a local constraint program, which is composed of:
\begin{itemize}
\item
  Sentences of factor $f$;
\item
  $l_{v^\prime \rightarrow f} \leq P\left( v^\prime \right) \leq u_{v^\prime \rightarrow f},\forall v^\prime\in N\left(f\right)\setminus\{v\}$;
\item
  Quadratic constraints that the variables in $N\left(f\right)\setminus\{v\}$ are independent of each other.
\end{itemize}
The objective function is $P\left( v \right)$, and the message $l_{f \rightarrow v}$ and $u_{f \rightarrow v}$ are the results of minimizing and maximizing the objective with the local constraint program.

In the example, the constraint program to update $l_{f_2 \rightarrow b}$ and $u_{f_2 \rightarrow b}$ is:
\begin{align}
  & 0.6 \leq P\left(b\mid a\right) \leq 0.7 \nonumber \\
  & 0.1 \leq P\left(b\mid \neg a \right) \leq 0.2 \nonumber \\
  & l_{a \rightarrow f_2} \leq P\left(a\right) \leq u_{a \rightarrow f_2} \nonumber\\
  & \textrm{minimize/maximize } P\left(b\right) \nonumber
\end{align}

For another example, suppose factor $f$ is one sentence of $0.3 \leq P\left( c \land \left( d \lor e \right)\right) \leq 0.4$, and its constraint program to update $l_{f \rightarrow d}$ and $u_{f \rightarrow d}$ would be:
\begin{align}
  & 0.3 \leq P\left( c \land \left( d \lor e \right)\right) \leq 0.4 \nonumber\\
  & l_{c \rightarrow f} \leq P\left(c\right) \leq u_{c \rightarrow f} \nonumber\\
  & l_{e \rightarrow f} \leq P\left(e\right) \leq u_{e \rightarrow f} \nonumber\\
  & P\left( c \land e \right) = P\left(c\right) \cdot P\left(e\right) \nonumber\\
  & \textrm{minimize/maximize } P\left(d\right) \nonumber
\end{align}

All $l$'s are initialized to 0 and $u$'s initialized to 1, and the messages are updated until convergence.
Intuitively, the role of factor nodes is to solve local constraint programs, and the role of variable nodes is to tighten the bounds.
The independence assumptions in the local constraint programs is a mechanism to approximate Markov condition, and the same approach is used in classical belief propagation.
There is a notable relation to the LNN inference algorithm \cite{lnn}, which also iteratively tightens bounds.
There is also a relation to the IPE algorithm \cite{l2u}: IPE cut out a number of polytree subgraphs, solve each subgraph, and then choose tightest bounds from the subgraphs; the new algorithm implicitly enumerates an exponential number of subgraphs and chooses one subgraph for each $l$ or $u$ that computes the tightest bound.
The algorithm guarantees correctness on polytree Bayesian and credal networks, with and without additional marginal probability sentences that break the unique-assessment assumption.
In general, however, there is no guarantee on correctness; this is the same as classical belief propagation.

Finally, let's make sure that the modified algorithm solves the example correctly.
The following are the messages at convergence.
\begin{align}
l_{f_1 \rightarrow a} = 0.2 & \,,\, u_{f_1 \rightarrow a} = 0.3 \nonumber\\
l_{a \rightarrow f_1} = 0.2 & \,,\, u_{a \rightarrow f_1} = 0.6 \nonumber\\
l_{a \rightarrow f_2} = 0.2 & \,,\, u_{a \rightarrow f_2} = 0.3 \nonumber\\
l_{f_2 \rightarrow a} = 0.2 & \,,\, u_{f_2 \rightarrow a} = 0.6 \nonumber\\
l_{f_2 \rightarrow b} = 0.2 & \,,\, u_{f_2 \rightarrow b} = 0.35 \nonumber\\
l_{b \rightarrow f_2} = 0.3 & \,,\, u_{b \rightarrow f_2} = 0.4 \nonumber\\
l_{b \rightarrow f_3} = 0.2 & \,,\, u_{b \rightarrow f_3} = 0.35 \nonumber\\
l_{f_3 \rightarrow b} = 0.3 & \,,\, u_{f_3 \rightarrow b} = 0.4 \nonumber
\end{align}
Therefore, we get the correct lower and upper bounds for $P\left(b\right)$:
\begin{align}
  l_b &= \max\left( l_{f_2 \rightarrow b}, l_{f_3 \rightarrow b} \right) = \max\left( 0.2, 0.3 \right) = 0.3 \nonumber\\
  u_b &= \min\left( u_{f_2 \rightarrow b}, u_{f_3 \rightarrow b} \right) = \min\left( 0.35, 0.4 \right) = 0.35 \nonumber
\end{align}

\section{Details of the Mastermind Experiments}

Algorithm~\ref{alg:mm} specifies how the puzzles are generated.
The reason that we run Knuth's algorithm three times is to obtain a longer board and thereby reduce the number of MAP ties that have the same posterior probabilities.
With three, most of the puzzles have a single MAP code to be used as the ground truth.
In the case that there are multiple MAP codes, we consider an inference algorithm correct if it guesses any one of them.

\begin{algorithm}[th]
    \caption{Generate a Mastermind puzzle.}
    \label{alg:mm}
    \begin{algorithmic}[1] % The number tells where the line numbering should start
        \State Sample a hidden code from uniform distribution.
        \State Sample each $P\left( l_i \right)$ uniformly from [0.3,0.7].
        \State Run Knuth's algorithm 3 times until success or contradiction. Each $l_i$ is sampled according to $P\left( l_i \right)$ from step 2. If $l_i=\text{True}$, uniformly sample a feedback from possible lies.
        \State If the hidden code is guessed in any of the 3 runs, exit.
        \State Save the board as a puzzle.
        \State Compute the exact MAP as the ground truth.
    \end{algorithmic}
\end{algorithm}

\begin{figure}[th]
  \centering
  \includegraphics[width=0.9\columnwidth]{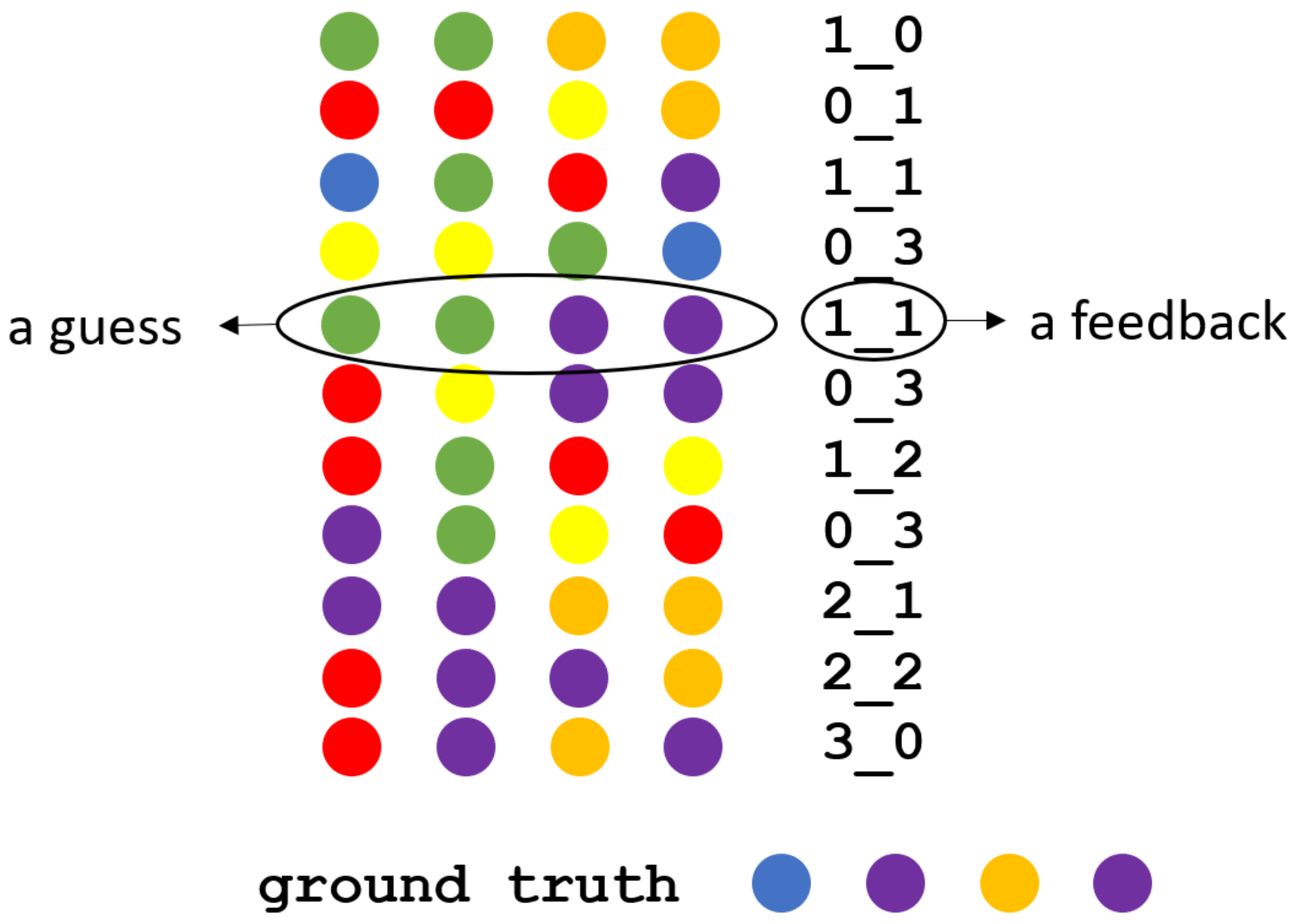}
  \caption{An example puzzle.}
  \label{fig:puzzle}
\end{figure}

Figure~\ref{fig:puzzle} illustrates one puzzle generated by Algorithm~\ref{alg:mm}.
Each row has two parts: the first is a guess by Knuth's algorithm, which is composed of 4 colored pegs out of six possible colors, the second part is the feedback which may or may not be a lie.

In addition to puzzles, we also need to generate knowledge as in (23-25) and alike.
As shown, the formulas are AND/OR of two consecutive $l_i$'s, and they alternate between AND and OR.
Note that we could replace them with arbitrary formulas.
For each formula, we first compute the exact point probability $p$ based on the $P\left( l_i \right)$ values from step 2 of Algorithm~\ref{alg:mm}.
Then we compute the widest probability interval $\left[x,y\right]$ for this formula if $P\left( l_i \right)$ could take any value in [0.3,0.7]: for AND, $x=0.09$ and $y=0.49$; for OR, $x=0.51$ and $y=0.91$.
Then we sample a number uniformly from $\left[x,p\right]$ as the lower bound for the formula, and sample a number uniformly from $\left[p,y\right]$ as the upper bound.
This ensures that the knowledge in sentences like (23-25) are correct.

\section{Details of the Credit Card Fraud Detection Experiments}

The LCN for credit card fraud detection task is the following.
Let $X$ denote the binary Is-Fraud variable; let $F_i$ denote the $i^\mathrm{th}$ feature variable in the Naive Bayes classifier, and let $f_{i,j}$ denotes the $j^\mathrm{th}$ possible value for $F_i$; let $c_{0,i,j}$ denote the probability of $F_i = f_{i,j}$ in legitimate transactions in training data, and let $c_{1,i,j}$ denote that for fraudulent ones; let $c_X$ denote the fraction of fraudulent transactions in training data; let $A_1,A_2,A_3$ denote the antecedents in the three logic rules.
\begin{align}
c_X     & \leq P\left( X \right) \leq c_X       \label{eq:sup1} \\
c_{0,i,j} & \leq P\left( F_i = f_{i,j}\mid\neg X \right) \leq c_{0,i,j} \,,\, \forall i,j \label{eq:sup2} \\
c_{1,i,j} & \leq P\left( F_i = f_{i,j}\mid X \right) \leq c_{1,i,j} \,,\, \forall i,j  \label{eq:sup3} \\
0.65    & \leq P\left( A'_1 \right) \leq 0.74 \\
0.31    & \leq P\left( A'_2 \right) \leq 0.66 \\
0.44    & \leq P\left( A'_3 \right) \leq 0.72 \\
1       & \leq P\left( X \mid A_1 \land A'_1 \right) \leq 1 \\
1       & \leq P\left( X \mid A_2 \land A'_2 \right) \leq 1 \\
1       & \leq P\left( X \mid A_3 \land A'_3 \right) \leq 1 
\end{align}
The first three equations (\ref{eq:sup1})(\ref{eq:sup2})(\ref{eq:sup3}) are exactly the Naive Bayes model.
The latter six equations use three auxiliary variables $A'_1,A'_2,A'_3$, and the effect is similar to the noisy-OR model in Bayesian\_midpoint and Credal.
However, unlike credal/Bayesian networks, LCN does not require unique-assessment assumption and therefore (\ref{eq:sup1}) is still allowed.
Table~2 shows that this flexibility of LCN results in substantial performance gains.